\documentclass[twoside,11pt]{article}

\usepackage{jmlr2e}
\usepackage{mathtools}
\usepackage{graphicx}
\usepackage{epsfig}
\usepackage{subfigure}

\newcommand{\dataset}{{\cal D}}

\ShortHeadings{Tutorial on Implied Posterior Probability for SVMs}{Nalbantov and Ivanov}
\firstpageno{1}

\begin{document}

\title{Tutorial on Implied Posterior Probability for SVMs}

\author{\name Georgi Nalbantov \email gnalbantov@mdscience.eu \\
	\name Svetoslav Ivanov \email sivanov@mdscience.eu \\
       \addr Department of Data Science, Medical Data Science Ltd., Bulgaria}

\editor{}
\today

\maketitle

\begin{abstract}
Implied posterior probability of a given model (say, Support Vector Machines (SVM)) at a point $\bf{x}$ is an estimate of the class posterior probability 
pertaining to the class of functions of the model applied to a given dataset. It can be regarded as a score (or estimate) for the true posterior probability, which can then be calibrated/mapped onto expected (non-implied by the model) posterior probability implied by the underlying functions, which have generated the data. In this tutorial we discuss how to compute
implied posterior probabilities of SVMs for the binary classification case as well as how to calibrate them via a standard method of isotonic regression.
\end{abstract}


\section{Introduction}
\label{sec:intro}

The implied posterior probability method for estimating class posterior probability has recently been proposed \citep{nalbantov2019note}. The method provides a score (or estimate) for the true posterior probability, which can then be calibrated/mapped onto expected (non-implied by the model) posterior probability implied by the underlying functions, which have generated the data. The main difference with other methods for solving this problem is the non-reliance on the original model built on the data to estimate posterior probabilities for points which do not belong to the separation surface of the model. Rather, the estimates are based on the class of functions used to build the (original) model, as applied to different versions of the dataset, where the relative weight of the instances varies between the classes. For each such relative weight a different model is built, which is relevant for the estimation of a particular value of the posterior probability. This value is derived from estimated ratio of class-probability density functions and the initial class priors via the Bayes formula for posterior probability.

Here we explore in greater detail the application of the method of implied posterior probability estimation for Support Vector Machines (SVMs). We consider the binary classification case, with instances of classes ``$+$'' and ``$-$'' from dataset $\dataset$. The (linear) SVM \citep{CortesVapnik95} approach to solving this task  involves funding the coefficients $\bf{w} $ and $b$ of the class-separating hyperplane which minimize the function

\begin{equation}
\label{eq:SVMformulation}
\frac{1}{2} \mathbf{w} \cdot \mathbf{w} + C \sum_{i=1}^{n} \xi_i
\end{equation}
s.t.
\[
y_i(\mathbf{x} \cdot \mathbf{w} + b) \geq 1 - \xi_i ,   i=1..n,
\]
\[
\xi_i \geq 0,   i=1..n,
\]

\noindent where $y$ is class label (``$+$'' or ``$-$''), $C$ is a manually set penalization parameter, $\xi$ is a slack (error) variable, and $n$ is the total number of points in the dataset. For concreteness, let us assume that $C = 5$, and we explicitly write separate $C$ for each class, that is $C = C^{+} = C^{-} = 5$. The optimization problem (\ref{eq:SVMformulation}) can thus be rewritten in the following form:

\begin{equation}
\label{eq:SVMformulation_C5}
\frac{1}{2} \mathbf{w} \cdot \mathbf{w} + C^{+} \sum_{i \in +} \xi_i + C^{-} \sum_{j \in -} \xi_j = 
\frac{1}{2} \mathbf{w} \cdot \mathbf{w} + 5 \sum_{i \in +} \xi_i + 5 \sum_{j \in -} \xi_j
\end{equation}
s.t.
\[
y_i(\mathbf{x} \cdot \mathbf{w} + b) \geq 1 - \xi_i ,   i=1..n^{+},
\]
\[
y_j(\mathbf{x} \cdot \mathbf{w} + b) \geq 1 - \xi_j ,   j=1..n^{-},
\]
\[
\xi_i \geq 0,   i=1..n,
\]

\noindent where $n^{+}$ and $n^{-}$ are the number of positive and negative instances in the dataset, respectively. Note that instead of interpreting the sum of errors $5 \sum_{i \in +} \xi_i$ as imposing $C^{+} = 5$ for each error $\xi_i$, we can equivalently state that we have $C^{+} = C^{-}= 1$ and 5 copies of each point from the dataset. In fact, if we have 5 times the initial dataset (each point occurring 5 times), then having the C parameter equal to 1 would result in the same solution of the optimization problem (\ref{eq:SVMformulation_C5}). The (equivalent to (\ref{eq:SVMformulation_C5})) formulation in this case is:

\begin{equation}
\label{eq:SVMformulation_C5_equiv}
\frac{1}{2} \mathbf{w} \cdot \mathbf{w} + 1 \sum_{i \in +} \xi_i + 1 \sum_{j \in -} \xi_j
\end{equation}
s.t.
\[
y_i(\mathbf{x} \cdot \mathbf{w} + b) \geq 1 - \xi_i ,   i=1..5n^{+},
\]
\[
y_i(\mathbf{x} \cdot \mathbf{w} + b) \geq 1 - \xi_i ,   j=1..5n^{-},
\]
\[
\xi_i \geq 0,   i=1..n.
\]

\noindent Therefore we refer to the value $C^{+}n^{+}$ as the effective number of class ``$+$'' points in the dataset, and $C^{-}n^{-}$ as the effective number of class ``$-$'' points in the dataset. The  effective total number of points in the dataset is thus $C^{-}n^{-} + C^{+}n^{+}$.

Changing the prior of a class in a dataset is achieved empirically by increasing the weight of each point from a given class with the same amount. For instance, if we double the amount of the ``$+$'' points, the function to be minimized becomes

\begin{equation}
\label{eq:SVMformulation_CplusDouble}
\frac{1}{2} \mathbf{w} \cdot \mathbf{w} + C^{+} \sum_{i \in +} \xi_i  + C^{+} \sum_{i \in +} \xi_i + C^{-}\sum_{j \in -} \xi_j =
 \end{equation}
\[
\frac{1}{2} \mathbf{w} \cdot \mathbf{w} + 2C^{+} \sum_{i \in +} \xi_i  + C^{-}\sum_{j \in -} \xi_j.
\]

\noindent In this case the effective total number of points in the dataset goes up to $C^{-}n^{-} + 2C^{+}n^{+}$. Interestingly, by altering $C^{+}$ we can increase the weight of the positive class continuously. For example, if we increase the weight of the positive class 1.5 times -- loosely speaking we add each point to the dataset 1.5 times --  then the function to be optimized becomes

\begin{equation}
\label{eq:SVMformulation_Cplus1.5}
\frac{1}{2} \mathbf{w} \cdot \mathbf{w} + 1.5C^{+} \sum_{i \in +} \xi_i  + C^{-}\sum_{j \in -} \xi_j,
\end{equation}

\noindent and the effective number of points would be $C^{-}n^{-} + 1.5 C^{+}n^{+}$.

Let us consider the case where we increase the weights of both classes equally. In this case we can write two equivalent SVM problem formulations (skipping the constraints):

\begin{equation}
\label{eq:SVMformulation_C+double_C-double}
\frac{1}{2} \mathbf{w} \cdot \mathbf{w} + 2C^{+} \sum_{i \in +} \xi_i  + 2C^{-}\sum_{j \in -} \xi_j,
\end{equation}

\noindent where $C^{+}$ and $C^{-}$ are increased two times, and

\begin{equation}
\label{eq:SVMformulation_2C+and_2C-}
\frac{1}{2} \mathbf{w} \cdot \mathbf{w} + C^{+} \sum_{i \in +} \xi_i  + C^{+} \sum_{i \in +} \xi_i + C^{-}\sum_{j \in -} \xi_j + C^{-}\sum_{j \in -} \xi_j,
\end{equation}

\noindent where the dataset has been doubled. Note that the total effective number of points in the dataset have doubled to $2C^{-}n^{-} + 2C^{+}n^{+}$. The relative proportion of points has remained equal however, as

\[
\frac{C^{+}n^{+}}{C^{-}n^{-}} = \frac{2C^{+}n^{+}}{2C^{-}n^{-}}.
\]

\noindent Although the relative proportion stays the same, the SVM separation hyperplane will be different since what we have effectively done is to increase the $C$ parameter 2 times, which generally results in deceased margin and changed position of the SVM class-separation surface. The change occurs due to the penalization term $0.5\bf{w}\cdot\bf{w}$. Without it the optimization problem would be the same, just rescaled by a constant (leaving aside the issue of the solution of such an optimization problem). For this reason from now on we consider the case where the effective total number of points in the dataset remains the same after altering $C^{+}$ and $C^{-}$. If we for instance increase $C^{+}$ by $\delta^{+}$, the resulting change in $C^{-}$ which ensures that the effective total number of points remains the same would be the solution for $\delta^{-}$ of the equation

\[
C^{+}n^{+} + C^{-}n^{-} = (C^{+} + \delta^{+}) n^{+} + (C^{-} - \delta^{-}) n^{-}
\]
\[
=>
\]
\[
\delta^{-} = \delta^{+} \frac{n^{+}}{n^{-}}.
\]

\noindent Thus, if $C^{+}$ is increased to $C^{+} + \delta^{+}$, in order to keep the effective total number points the same, we would need to decrease $C^{-}$ to $C^{-} - \delta^{-} = C^{-} - \delta^{+} n^{+}/n^{-}$. One condition to consider is that $C^{-} - \delta^{-} > 0$, implying that we cannot increase the effective total number of points from one class indefinitely, since there is a constraint on the total effective number of points (which should stay the same).

The reason we stress the importance of keeping the effective total number of points in the dataset is the ability to change the ratio of priors of the two classes from

\[
\frac{C^{+}n^{+}}{C^{-}n^{-}} \text{ \hspace{0.1cm}  to  \hspace{0.1cm}  } \frac{(C^{+} + \delta^{+}) n^{+}}{C^{-} - \delta^{+} \frac{n^{+}}{n^{-}}},
\]

\noindent while not distorting the total number of points in the dataset. This is useful in optimization formulations of type ``loss + penalty'' , whereas in the absence of a penalty term there is no effect on the final solution as the loss (cost) function being optimized is just rescaled by a parameter, which does not affect the values of the estimated unknown coefficients.  

\section{Estimation of class-posterior probabilities}
\label{sec:estimation_postprob}

\begin{figure*}[h]
 	\center
 	\hspace*{-0.85cm}
 		\includegraphics[angle=0,origin=c,height=5cm]{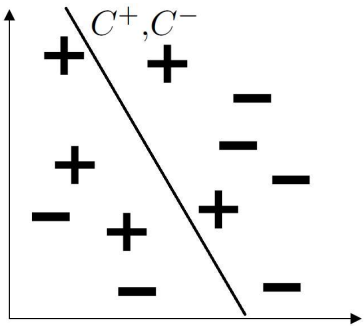}
 	\vspace*{-0.5cm}
 	\caption{A two-class dataset with SVM separation line (for some $C^{+}$,$C^{-}$).}
	\label{fig:SVM_dataset}
 \end{figure*}

Now that we know how to change the relative weight of (estimated) class priors without changing the total effective number of points in the dataset, we can move to the estimation of implied class-posterior probabilities for SVMs. The setup is as follows. Consider the 2D binary classification case in Figure~\ref{fig:SVM_dataset}, with the SVM separation line drawn. 

We would like to estimate the posterior probability of any point $\bf{x}$ in the domain of the dataset, which is suggested (implied) by the SVM model's class of functions. Such a posterior probability is the estimate (score) of the true posterior probability. It need not be calibrated by construction, as the model may not approximate well the underlying function which has generated the dataset. The only place where we can currently estimate the posterior class probabilities is along the SVM separation line, where $\widehat{P\left(+ \mid \bf{x} \right)} = \widehat{P\left(- \mid \bf{x} \right)} = 0.5$, which is trivial. From the Bayes formula 

\[
P\left( - \mid \bf{x} \right)  = \frac{f\left( \bf{x} \mid + \right) P\left( + \right)}{P\left( \bf{x} \right)}
\]

\noindent we can derive that in this case

\[
\frac{0.5}{0.5} = \frac{P\left(+ \mid \bf{x} \right)}{P\left(- \mid \bf{x} \right)}  = \frac{f\left( \bf{x} \mid + \right) P\left( + \right)}{f\left( \bf{x} \mid - \right) P\left( - \right)} = 1
\]
\[
=>
\]
\[
\frac{f\left( \bf{x} \mid + \right)}{f\left( \bf{x} \mid - \right)} = \frac{P\left( - \right)}{P\left( + \right)}
\]

\noindent along the true separation surface. The ratio of class priors is estimated by 

\[
\frac{\widehat{P\left( - \right)}}{\widehat{P\left( + \right)}} = \frac{C^{-}n^{-}}{C^{+}n^{+}}
\]

\noindent in terms of the total effective number of points from both classes. If we increase the weight of ``$+$'' points to $C^{+} + \delta^{+}$ and simultaneously decrease the weight of  ``$-$'' points to $C^{-} - \delta^{+}n^{+}/n^{-}$ we will end up with an estimated relative ratio of priors as 

\[
 \frac{(C^{-} - \delta^{+}\frac{n^{+}}{n^{-}})n^{-}}{(C^{+} + \delta^{+})n^{+}}.
\]

\begin{figure*}[h]
 	\center
 	\hspace*{-0.85cm}
 		\includegraphics[angle=0,origin=c,height=10cm]{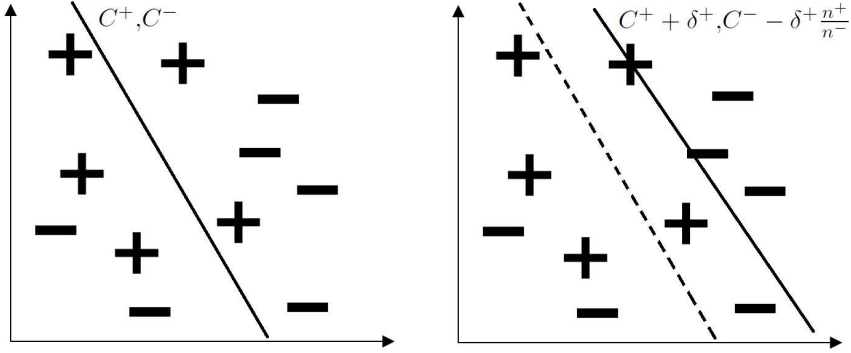}
 	\vspace*{-4.5cm}
 	\caption{A shift of the SVM separation line (for some $C^{+}+\delta^{+}$, $C^{-}-\delta^{+}\frac{n^{+}}{n^{-}}$).}
	\label{fig:SVM_dataset_line_shifted} 
\end{figure*}

\noindent If we build an SVM model with weights $C^{-} - \delta^{+}\frac{n^{+}}{n^{-}}$ for the ``$-$'' class and $C^{+} + \delta^{+}$ for the ``$+$' class the function to be minimized would be

\begin{equation}
\label{eq:SVMformulation_C+delta_and_C-delta}
\frac{1}{2} \mathbf{w} \cdot \mathbf{w} + (C^{+} + \delta^{+})\sum_{i \in +} \xi_i  + (C^{-} - \delta^{+}\frac{n^{+}}{n^{-}})\sum_{j \in -} \xi_j 
\end{equation}

\noindent and we will end up with a different location of the SVM separation hyperplane (see Figure~\ref{fig:SVM_dataset_line_shifted}, right panel).

For all points $\bf{x}$ along this new separation surface, which we denote by hyperplane $h$ obtained using parameters $\delta^{+},C^{+},\delta^{-},C^{-}$ on dataset $\dataset$, that is,

\[
\forall \mathbf{x} \in h(\delta^{+},C^{+},\delta^{-},C^{-},\dataset) ,
\]

\noindent where $\delta{-} = \delta^{+}n^{+}/n^{-}$, we can estimate that the ratio of class probability density functions equals the (inverse) ratio of the estimated class prior probabilities:

\[
\widehat{\left(\frac{f\left( \bf{x} \mid + \right)}{f\left( \bf{x} \mid - \right)}\right)}  = \frac{(C^{-} - \delta^{+}\frac{n^{+}}{n^{-}})n^{-}}{(C^{+} + \delta^{+})n^{+}},
\]

\noindent where the \emph{hat} is over the ratio of the class-conditional densities and $\bf{x}$ lies on the SVM hyperplane (the separation surface) resulting from solving Eq.(\ref{eq:SVMformulation_C+delta_and_C-delta}). In other words, we have just estimated the ratio of class probability density functions at a point $\bf{x}$. Having this ratio is enough to compute the class posterior probabilities at point $\bf{x}$ implied by the SVM model's function class, as according to the Bayes formula

\begin{equation}
\label{eq:pnew_ff_pp}
P\left(+ \mid \bf{x} \right) = 1/\left({1+  \frac{f\left( \bf{x} \mid + \right) P\left( + \right)}
{f\left( \bf{x} \mid - \right) P\left( - \right)}}\right)
\end{equation}

\noindent for any point $\bf{x}$. The ratio $P\left( + \right) / P\left( - \right)$ is approximated by $C^{+} n^{+} / C^{-} n^{-}$, and the ratio $f\left( \bf{x} \mid + \right) / f\left( \bf{x} \mid - \right)$ is approximated by $((C^{-} - \delta^{+}\frac{n^{+}}{n^{-}})n^{-}) / ((C^{+} + \delta^{+})n^{+})$, or:

\begin{equation}
\label{eq:postptob_hat}
\widehat{P\left(+ \mid \bf{x} \right)} = 1/\left(1+  \frac{(C^{-} - \delta^{+}\frac{n^{+}}{n^{-}})n^{-}}{(C^{+} + \delta^{+})n^{+}} \cdot \frac{C^{+} n^{+} }{C^{-} n^{-}} \right) ~ ~ ~ \forall \mathbf{x} \in h(\delta^{+},C^{+},\delta^{-},C^{-},\dataset)
\end{equation}

We note the importance of increasing (decreasing) the weight of a class by increasing (decreasing) the weight of each point from the class equally. In this way the (empirical) probability density function of the class remains the same. This ensures that the (empirical) ratio of probability densities of the classes at any point $\bf{x}$ stays the same when the prior probabilities  change due to the change in class weights. 

In the case of a balanced dataset ($n^{+} = n^{-}$) and equal $C$ constant for both classes, the above equation reduces to 

\[
\widehat{P\left(+ \mid \bf{x} \right)} = 0.5 + 0.5 \frac{\delta^{+}}{C}
\]

Another special case arises when $C^{+}n^{+} = C^{-}n^{-}$, the penalization parameter of the positive class is defined as $0.5C^{+}$, and the penalization parameter of the negative class is defined as $0.5C^{-}$. In this case initially the total effective number of points $n$ in the dataset is

\[
n = 0.5C^{+}n^{+}  + 0.5C^{-}n^{-}.
\]

\noindent The penalization parameter for the positive class is increased to $z^{+}C^{+}$ and the penalization parameter for the negative class is reduced to $z^{-}C^{-}$. To ensure that the total effective number of points remains constant, we need

\[
z^{+}C^{+}n^{+}  + z^{-}C^{-}n^{-} = 0.5C^{+}n^{+}  + 0.5C^{-}n^{-}
\]
\[
=>
\]
\[
z^{-} = 0.5 + \frac{(0.5 - z^{+})C^{+}n^{+}}{C^{-}n^{-}} = 1 - z^{+}.
\]

\noindent The last equality holds due to the assumption of $C^{+}n^{+} = C^{-}n^{-}$. Therefore

\[
\widehat{\left(\frac{f\left( \bf{x} \mid + \right)}{f\left( \bf{x} \mid - \right)}\right)}  = \frac{z^{-}C^{-}n^{-}}{(1-z^{-})C^{+}n^{+}},
\]

\noindent and

\begin{equation}
\label{eq:postptob_hat_equaleffective}
\widehat{P\left(+ \mid \bf{x} \right)} = 1/\left(1+  \frac{z^{-}C^{-}n^{-}}{(1-z^{-})C^{+}n^{+}} \cdot \frac{0.5C^{+} n^{+} }{0.5C^{-} n^{-}} \right) = 1 - z^{-} = z^{+}  ~ ~ ~ \forall \mathbf{x} \in h(z^{+},C^{+},z^{-},C^{-},\dataset).
\end{equation}

\noindent That is, we end up with a simple estimate for the positive-class posterior probability, which is equal to the percentage weight of the positive class in the (re-weighted) dataset, $z^{+}$. 

In the most general case, without imposing restrictions on $C^{+}$, $C^{-}$, $n^{+}$, and $n^{-}$, the estimated posterior probability (for the positive class) in case the penalization constant for the positive instances is $0.5C^{+}$ (implying $z^{+}=0.5$) and the penalization constant for the negative instances is $0.5C^{-}$ (implying $z^{-}=0.5$) is equal to:

\begin{equation}
\label{eq:postptob_hat_general}
\widehat{P\left(+ \mid \bf{x} \right)} = 1/\left(1+  \frac{ (0.5 + \frac{(0.5 - z^{+})C^{+}n^{+}}{C^{-}n^{-}}) C^{-}n^{-}}{z^{+}C^{+}n^{+}} \cdot \frac{0.5C^{+} n^{+} }{0.5C^{-} n^{-}} \right) =
\end{equation}
\[
\frac{z^{+}}{z^{+} + 0.5 + 0.5 \frac{C^{+}n^{+}}{C^{-}n^{-}} - z^{+}  \frac{C^{+}n^{+}}{C^{-}n^{-}}} = 
\frac{z^{+}C^{-}n^{-}}{(0.5 + z^{+}) C^{-}n^{-} + (0.5 - z^{+})C^{+}n^{+}}
\]
\[
\forall \mathbf{x} \in h(z^{+},C^{+},z^{-},C^{-},\dataset)
\]

\noindent over points $\bf{x}$ of the separation surface (hyperplane) built using penalization parameters  $z^{+}C^{+}$ for the positive class and  $z^{-}C^{-}$ for the negative class, where $z^{-} = 0.5 + (0.5 - z^{+})C^{+}n^{+}/(C^{-}n^{-})$. The conditions on $z^{+}$ and $z^{-}$ are (derived from the condition to keep the total effective number of points in the dataset constant):

\[
0 < z^{+} < \frac{0.5C^{+}n^{+} + 0.5C^{-}n^{-}}{C^{+}n^{+}}
\]
\[
0 < z^{-} <\frac{0.5C^{+}n^{+} + 0.5C^{-}n^{-}}{C^{-}n^{-}},
\]

\noindent bearing in mind that the function SVMs minimize in this case is

\[
\frac{1}{2} \mathbf{w} \cdot \mathbf{w} + z^{+} C^{+} \sum_{i \in +} \xi_i  + z^{-} C^{-}\sum_{j \in -} \xi_j
\]

\noindent as well as that

\[
z^{-} = 0.5 + \frac{(0.5 - z^{+})C^{+}n^{+}}{C^{-}n^{-}}.
\]

\section{Discussion and degeneracies}
\label{sec:discussion}

We have so far considered a method for estimation of posterior class probabilities (for the binary classification case). The method provides an estimate for these posterior probabilities, which can be regarded as a score that can further be calibrated to match true posterior probabilities by applying, for example, the method of isotonic regression as in \citep{Zadrozny2002}.

 \begin{figure*}[h]
 	\center
 		\includegraphics[height=5cm]{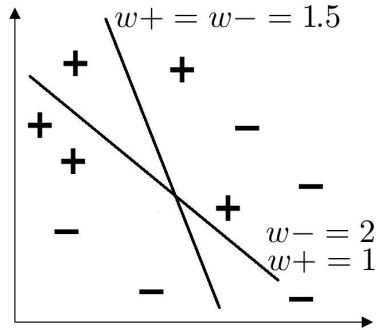}
 	\vspace*{0cm}
 	\caption{Illustration of a degeneracy. A classification algorithm produces two separation surfaces that cross each other when the relative weights of the positive $w+$ and negative $w-$ instances are changed.}
 		\label{fig:fig_deg}
 \end{figure*}

Before calibrating the estimates for class-posterior probabilities however let us first discuss one degeneracy, which could stem from the SVM modeling algorithm. Since there is no guarantee (and there is not expected to be one) that a change in the relative weight of the $C^{+}$ and $C^{-}$ would result in an SVM separation hyperplane that is parallel to the initial SVM hyperplane, there would be potentially more than one estimate (score) for the class-posterior probability at the crossing point of the hyperplanes. That is, if point $\bf{x}$ lies on the hyperplane computed using $C^{+}$ and $C^{+}$, as well as it lies on the hyperplane computed using $z^{+}C^{+}$ and $z^{-}C^{+}$, where $z^{+}$ and $z^{-}$ have been chosen such that the total effective number of instances in the dataset remains fixed, then we would have two different estimates for the class-posterior probability. This is a degeneracy arguably ``by construction'', as SVM algorithm permits one and the same estimate to originate from two different class priors. We expect this degeneracy to occur for all type of classification methods, which employ penalization type of parameters in their estimation procedures. In this case we end of with two (or potentially even more) ``scores'' for the class-posterior probability as a given point, which is not a favorable property. Such a degeneracy is illustrated in Figure \ref{fig:fig_deg}. In this figure, two SVM lines are drawn. The first line is calculated with weights equal to $0.5C^{+} = 0.5C^{-} \equiv w = 1.5$ (corresponding to parameter values $C^{+} = C^{-} = 3$). The second line is calculated with weights on the classes equal to $0.5 \times (2/3) \times C^{+} \equiv w^{+} = 1$ for the positive class and $0.5 \times  (4/3)  \times C^{-} \equiv w^{-} = 2$ for the negative class. The effective total number of points in the dataset in both cases is the same, and is equal to 15: $C^{+}n^{+} + C^{-}n^{-} = 1.5 \times 5  + 1.5 \times 5 = 15$ (for the first line), and $C^{+}n^{+} + C^{-}n^{-} = 1 \times 5  + 2 \times 5 = 15$ (for the second line), bearing in mind that the number of points for each class is equal to 5. Nevertheless, both lines intersect. At the intersection point we would compute 2 scores (estimates) for the posterior class probability. For the first line, at the crossing point $\bf{x}$ this estimate is:

\[
\widehat{P\left(+ \mid \bf{x} \right)} = 1/\left(1+  \frac{z^{-}C^{-}n^{-}}{(1-z^{-})C^{+}n^{+}} \cdot \frac{0.5C^{+} n^{+} }{0.5C^{-} n^{-}} \right) = z^{+} = 0.5,
\]
\[
\forall \mathbf{x} \in h(z^{+}=1.5,C^{+},z^{-}=1.5,C^{-},\dataset),
\]

\noindent where $h(z^{+},C^{+},z^{-},C^{-},\dataset)$ is the SVM separation hyperplane obtained on dataset $\dataset$ using parameters $z^{+},C^{+},z^{-},C^{-}$. For the second line, at the crossing point $\bf{x}$ this estimate is:

\[
\widehat{P\left(+ \mid \bf{x} \right)} = 1/\left(1+  \frac{z^{-}C^{-}n^{-}}{(1-z^{-})C^{+}n^{+}} \cdot \frac{0.5C^{+} n^{+} }{0.5C^{-} n^{-}} \right) = z^{+} = 0.5  \times \frac{2}{3} = \frac{1}{3},
\]
\[
\forall \mathbf{x} \in h(z^{+}=1,C^{+},z^{-}=2,C^{-},\dataset).
\]

\noindent Thus, we end up with two estimates at point $\bf{x}$ for the (positive) class posterior probability, which constitutes a degeneracy. We propose to tackle this case in the following way. Instead of trying to find one hyperplane what goes through a point $\bf{x}$, we will build many hyperplanes with varying relative weights for the classes (keeping the effective total number of points fixed). To be more precise, we consider the set $\cal H$ of all hyperplanes $h(z^{+},z^{-},C^{+},C^{-},\dataset)$ obtained from different pairs $(z^{+}, z^{-})$, which yield posterior class probability estimates that are uniformly distributed in the interval (0,1), while keeping the effective total number of points in the dataset the same. That is:

\[
{\cal H} \coloneqq \{ h (z^{+},z^{-} ) : \widehat{P\left(+ \mid \bf{x} \right)}\mid_{\mathbf{x} \in h({z}+,z-)} \sim  U(0,1)\},
\]
\noindent where pairs $(z^{+}, z^{-})$, which are different for each hyperplane $h$, satisfy the condition of $
z^{+}C^{+}n^{+} + z^{-}C^{-}n^{-} = $ the total effective number of points in the dataset.

 In a practical situation, we may consider a set of 99 hyperplanes, along which the estimated posterior probabilities are 0.01, 0.02, .. 0.99 (and add 2 more fictitious hyperplanes along which we define the posterior probabilities to be 0 and 1). For each hyperplane we can compute whether the estimated class for $\bf{x}$ is positive or negative (or, by convention no class if the point lies on the SVM separation hyperplane \footnote{In case $\bf{x}$ lies on a separation hyperplane, then we have found (at least) one estimate for the posterior probability, equal to $z^{+}$, where the parameters used to build the hyperplane are $z^{+}, z^{-}, C^{+}, C^{-}, n^{+}, \text{and } n^{+}$, as described in the procedures up till now.}). Equally important, for each relative weight combination of the classes for which the resulting hyperplane does not contain point $\bf{x}$, we have an estimate for whether at point $\bf{x}$ the posterior probability is less than or more than the posterior probability pertaining to the respective hyperplane (coming from the fact that we know whether each hyperplane predicts point $\bf{x}$ as positive or negative class). Thus, we can calculate the expected value of all these estimates, defining them to be 0 in case of predicted class being negative, and 1 is case of predicted class being positive (and, trivially, 0.5 in case point $\bf{x}$ lies on a hyperplane). We note that this estimation strategy is consistent with the case when there is no degeneracy: if there is only one hyperplane which goes via point $\bf{x}$, and say, the posterior probability is estimated as 0.3, then 30\% of the hyperplanes from the set $\cal H$ of hyperplanes which yield a uniform distribution of the posterior probabilities are bound to classify this point as positive, and 70\% as negative. The expected class label over this set of hyperplanes is $0.3 \times 1 + 0.7 \times 0 = 0.3$ (assuming the positive class is denoted by ``+1'', and the negative as ``0''), and so is the estimated class posterior probability. 


In light of this, we define the final estimate of the posterior probability (for the positive class) to be:

\begin{equation}
\label{eq:postptob_combined}
\widehat{P\left(+ \mid \bf{x} \right)} = E \left( cl(h,\mathbf{x}) \right),
\end{equation}



\noindent where the expectation is taken with respect to all hyperplanes from $\cal H$ ,
$cl(h(z^{+},C^{+},n^{+},C^{-},n^{-}),\bf{x})$ is the classification label $(\in ({0,1})$ for the negative and positive class, respectively) output by model (or, hyperplane) $h$, and we define $h(z^{+}=1,C^{+},n^{+},C^{-},n^{-}) \equiv 1$, and $h(z^{+}=0,C^{+},n^{+},C^{-},n^{-}) \equiv 0$. This estimate provides a single score at point $\bf{x}$, which does not suffer from the degeneracy problem outlined above. This is a de facto majority voting classification rule where all models (hyperplanes) from the set yielding uniformly distributed posterior probability estimates have the same classification weight. 
%

One point of discussion is whether to put greater weight on estimates, which are close to the extremes of 0 and 1, which we avoid currently and leave it to future research.

Another point of discussion, which is intrinsic to SVMs, and which we do not pursue in detail here, is the question of whether to keep only the support vectors from the original dataset and consider them only when applying different weights to the instances of the classes in the class-posterior estimation procedure. The points to consider here are that the ratio of support vectors of both classes may not generally equal to the ratio of number of instances of both classes, and that changing the weights $C^{+}$ and $C^{-}$ in the SVM optimization problem would result in a change in the number and type of support vectors. We propose to keep the initial dataset, and not discard any points in the class-posterior estimation procedure. This comes in contrast to estimation procedures based on scores produced by the initial SVM separation hyperplane, as for example proposed by Platt \citep{Platt99probabilisticoutputs}.

The proposed method for estimation of class-posterior probabilities does not consider in any way any ``score'' produced by the separation surface of the (initial) model built on the data. As such, it is quite different from the majority of methods, which rely on such scores. Nevertheless, these scores and the estimates of the method of implied posterior probabilities yield the same results over a wide variety of classification techniques (which do not have a penalization type of parameters in their formulation). We also stress that the proposed method is non-parametric in the sense that it does not try to approximate the parameters of a (parametric) distribution which might have generated the dataset.

Often the same penalization is used for both classes, that is $C^{+}=C^{-}$, even though a dataset may be unbalanced ($n^{+} \neq n^{-}$). In such a case, one tends to assume that the resulting separation line corresponds to the $\widehat{P\left(+ \mid \bf{x} \right)} = 0.5$ level. In such cases we propose to apply the class re-weighting scheme relevant for the case when the effective number of points from both classes is the same (even though it is in fact not the same), and apply Eq. (\ref{eq:postptob_hat_equaleffective}) for the estimation of posterior probabilities.

\section{A complete worked-out example on a 2D dataset}
\label{sec:2Ddataset}

Consider an artificially created two-class dataset in 2D, where the number of positive points is $n^{+} = 10$, and the number of negative points is $n^{-} = 10$. For each class, the data has been generated by a Gaussian with a class-specific mean vector and variance-covariance matrix. The penalization parameters for the classes are $C^{+} = C^{-} = 20$. In this case $C^{+}n^{+} = C^{-}n^{-}$, therefore we choose for convenience the SVM objective function to be

\[
\frac{1}{2} \mathbf{w} \cdot \mathbf{w} + 0.5C^{+} \sum_{i \in +} \xi_i + 0.5C^{-} \sum_{j \in -} \xi_j.
\]

Note that in this formulation the effective penalization becomes $0.5 \times 20 = 10$. Therefore, if we would like to have the effective penalization equal to 20, we should have chosen $C^{+} = C^{-} = 40$.

\begin{figure*}[h]
 	\center
	\hbox{\hspace{-2.5cm} \includegraphics[height=15cm]{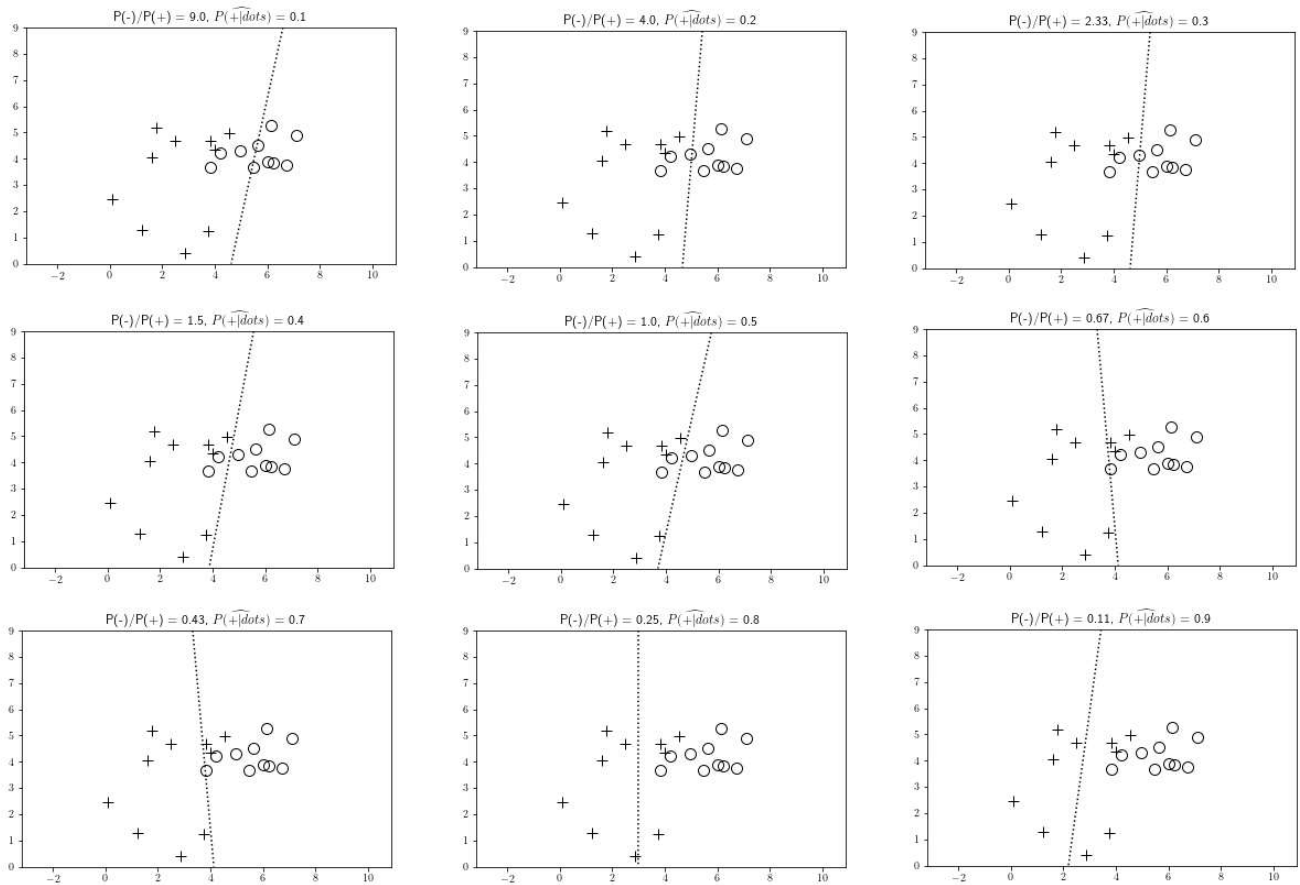}}
 	\vspace*{-2cm}
 	\caption{Nine SVM hyperplanes along which the posterior positive-class probabilities are estimated to be $0.1, 0.2, \dots ,0.9$.  Note that the hyperplanes are not (and need not be) parallel.}
 		\label{fig:9hp}
 \end{figure*}

The steps for computing the posterior probability at point $\bf{x}$ run as follows:

\medskip

Step 1. Build 9 hyperplanes \footnote{In a practical case, we would like to build more hyperplanes, e.g. 20 or 100, in order to have a more precise final estimate of the posterior probability.} corresponding to 9 equally-spaced values for the positive-class posterior probabilities in the interval [0.1,0.9]. These values are $0.1, 0.2, \dots ,0.9$. Thus, along the first hyperplane this posterior probability is estimated to be 0.1, along the second hyperplane this posterior probability is estimated to be 0.2, and so on until the last, $9^{th}$ hyperplane. These nine hyperplanes are shown in Figure \ref{fig:9hp}. Note that the 9 hyperplanes are not parallel, which implies that for some points in the domain of the dataset a degeneracy will occur. The way to build the hyperplane, along which the estimated (positive-class) posterior probability is 0.1, taken as an example, is derived from the estimate for the posterior probability in Eq.(\ref{eq:postptob_hat_equaleffective}) as being equal to $z^{+}$, which can be used as shorthand for Eq.(\ref{eq:postptob_hat_general}) since $C^{+}n^{+} = C^{-}n^{-}$:

\[
\widehat{P\left(+ \mid \bf{x} \right)} = 1/\left(1+  \frac{z^{-}C^{-}n^{-}}{(1-z^{-})C^{+}n^{+}} \cdot \frac{0.5C^{+} n^{+} }{0.5C^{-} n^{-}} \right) = 
\]

\[
\widehat{P\left(+ \mid \bf{x} \right)} = 1/\left(1+  \frac{0.9C^{-}n^{-}}{0.1C^{+}n^{+}} \cdot \frac{0.5C^{+} n^{+} }{0.5C^{-} n^{-}} \right) = z^{+} = 0.1.
\]

\noindent Therefore, the hyperplane along which the positive-class posterior probability is estimated to be 0.1 is built using the following SVM objective function:

\[
\frac{1}{2} \mathbf{w} \cdot \mathbf{w} + 0.1C^{+} \sum_{i \in +} \xi_i + 0.9C^{-} \sum_{j \in -} \xi_j.
\]

\noindent That is, in our case having $0.1 C^{+}$ as penalization parameter for the positive class, and $0.9 C^{-}$ as penalization parameter for the negative class ensures that the posterior probability estimate along the resulting hyperplane is 0.1. The rest (eight) of the hyperplanes are built analogically, using pairs of weights $0.2 C^{+}$ and $0.8 C^{-}$ for the second hyperplane (along which the positive-class posterior probability is estimated to be 0.2), $0.3 C^{+}$ and $0.7 C^{-}$ for the third hyperplane (along which the positive-class posterior probability is estimated to be 0.3) and so on. We note that two more (fictitious) hyperplanes are constructed where the penalization parameters for the positive class are $0C^{+}$ and $1C^{+}$, with corresponding posterior probability predictions of 0 and 1. 

\medskip

\begin{figure*}[h]
 	\center
 		\includegraphics[height=6cm]{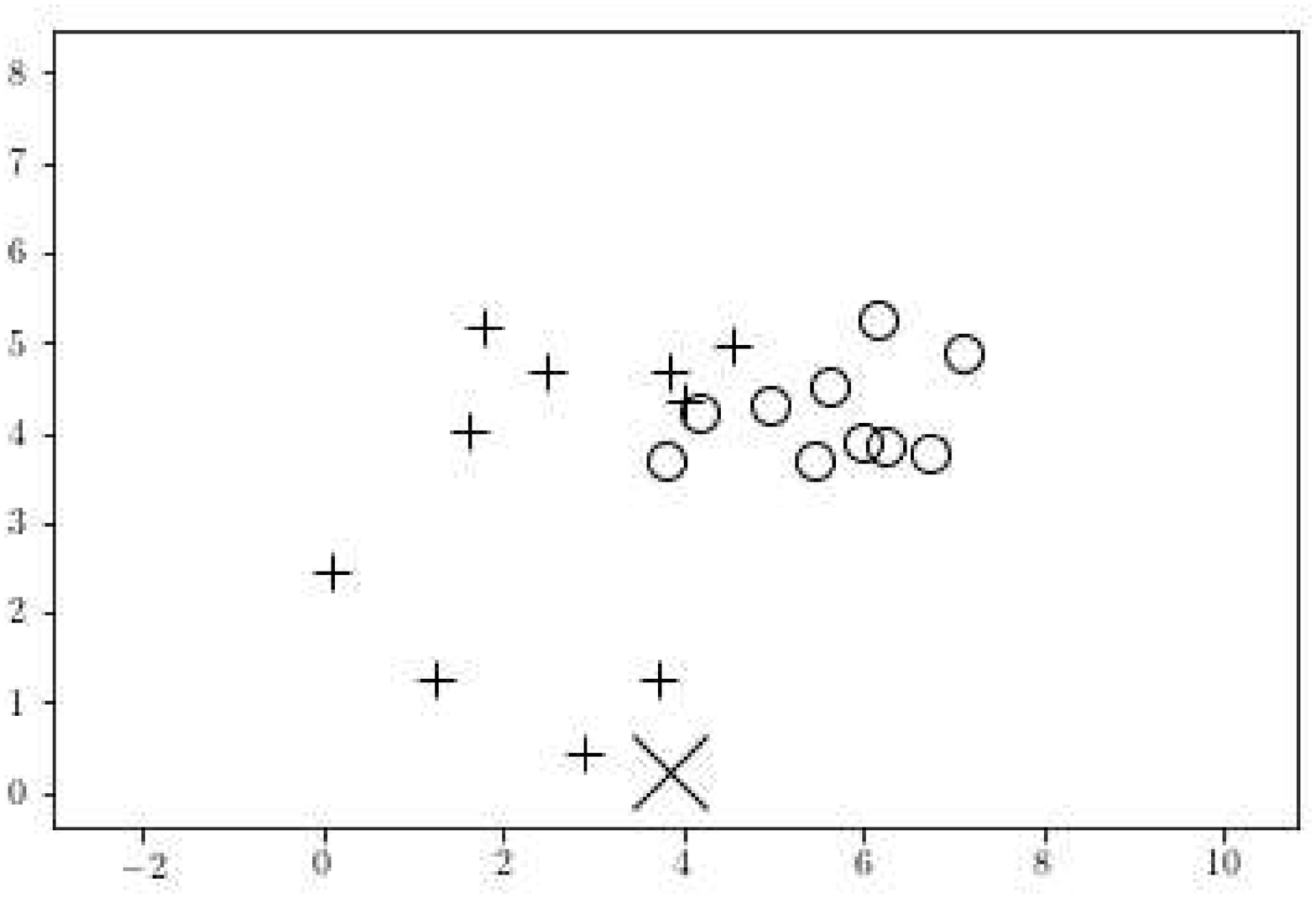}
 	\vspace*{0cm}
 	\caption{A 2D binary dataset and a test point \bf{x}. }
 		\label{fig:2D_ponitX}
 \end{figure*}

Step 2. Pick a test point $\bf{x}$ for which the posterior probability must be computed, as in Figure \ref{fig:2D_ponitX}. 

\medskip

\begin{figure*}[h]
 	\center
 		\includegraphics[height=6cm]{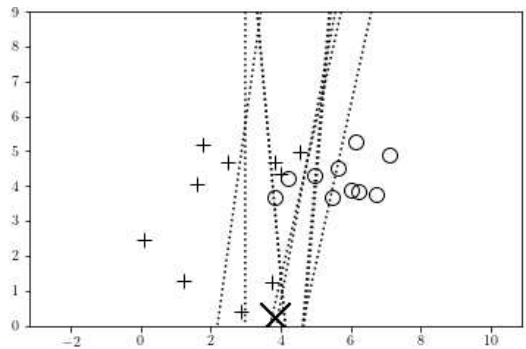}
 	\vspace*{-1cm}
 	\caption{A 2D binary dataset and a test point $\bf{x}$, together with nine SVM hyperplanes (from Figure \ref{fig:9hp}) along which the positive class posterior probabilities are $0.1, 0.2, \dots , 0.9$.}
 		\label{fig:9hp_pointX}
 \end{figure*}

\begin{figure*}[h]
 	\center
	\hbox{\hspace{-0.95cm} \includegraphics[height=13cm]{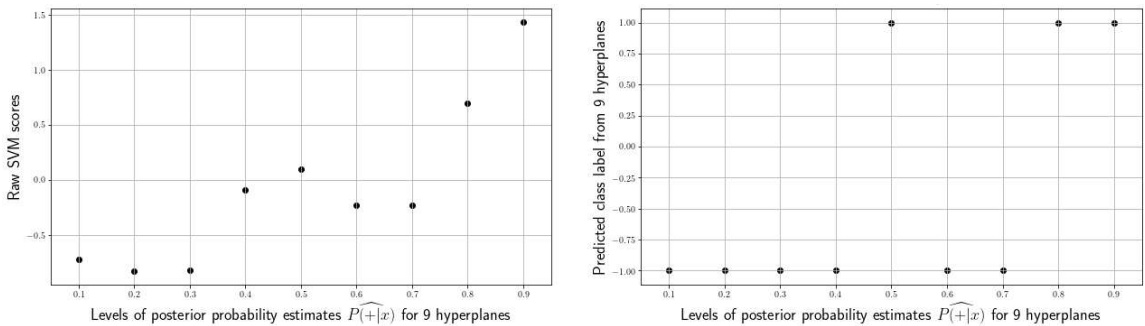}}
 	\vspace*{-7.5cm}
 	\caption{Raw SVM scores (left) and the corresponding predicted class labels at cutoff=0 (right) of the nine hyperplanes for point $\bf{x}$ from Figure \ref{fig:9hp_pointX}. The final estimated positive-class posterior probability for point $\bf{x}$ is calculated as the relative number of positive-class classifications, in this case $(3+1)/(9+2) = 4/11$.}
 		\label{fig:postProb_rawScore_predClass}
 \end{figure*}

Step 3. Classify point $\bf{x}$ with each of the 9 hyperplanes. In this concrete case, the 9 hyperplanes, which are the same as in Figure \ref{fig:9hp}, are shown in Figure \ref{fig:9hp_pointX}. We do not need to know the 9 raw SVM scores explicitly, just the predicted classifications (in this case, the sign of the raw scores) at $\bf{x}$. Out of the 9 classifications, 6 are negative (hyperplanes corresponding to expected posterior probabilities of 0.1, 0.2, 0.3, 0.4, 0.6, and 0.7) and 3 are positive (hyperplanes corresponding to expected posterior probabilities of 0.5, 0.8, 0.9) - see Figure \ref{fig:postProb_rawScore_predClass}. We note that point $\bf{x}$ has been chosen in such a way that a degeneracy to occur, since the $5^{th}$ hyperplane classifies $\bf{x}$ as positive, while the $4^{th}$ and the $6^{th}$ classify $\bf{x}$ as negative. 

\medskip

Step 4. Compute the (final) estimate for the positive-class posterior probability as the percentage of hyperplanes that classify $\bf{x}$ as positive, in this case $4/11$. We note that this estimate is not $3/9$, as we have added two fictitious hyperplanes (one of which classifies point $\bf{x}$ as positive, and the other as negative, whatever the test point).


\medskip

\begin{figure*}[h]
 	\center
 		\includegraphics[height=6cm]{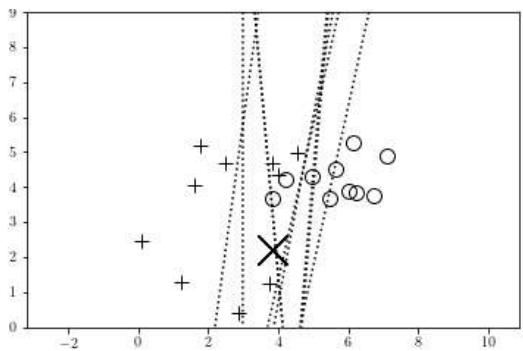}
 	\vspace*{-1cm}
 	\caption{A 2D binary dataset and a text point $\bf{x}$, together with nine SVM hyperplanes (from Figure \ref{fig:9hp}) along which the positive class posterior probabilities are $0.1, 0.2, \dots ,0.9$.}
 		\label{fig:9hp_pointX_noDeg}
 \end{figure*}

\begin{figure*}[h]
 	\center
		\hbox{\hspace{-0.95cm} \includegraphics[height=13cm]{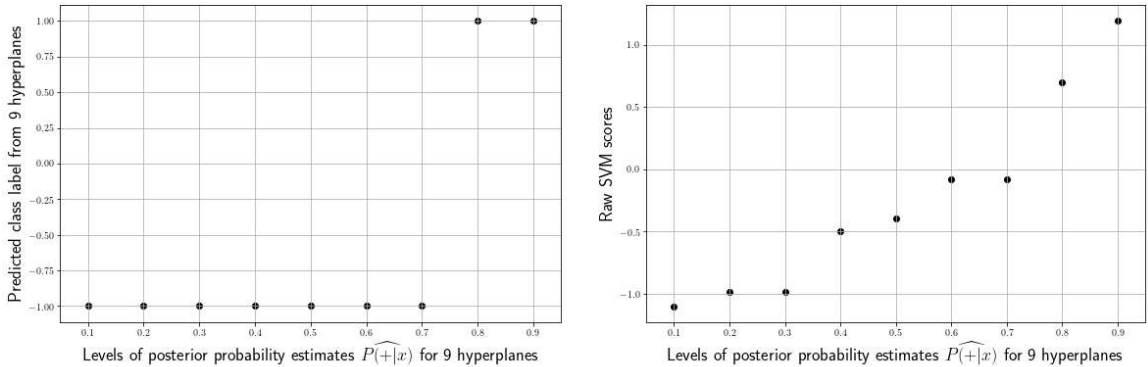}}
 	\vspace*{-7cm}
 	\caption{Raw SVM scores (left) and the corresponding predicted class labels at cutoff=0 (right) of the nine hyperplanes for point $\bf{x}$ from Figure \ref{fig:9hp_pointX_noDeg}. The estimated positive-class posterior probability for point $\bf{x}$ is estimated as the relative number of positive-class classifications, in this case (2+1)/(9+2) = 3/11.}
 		\label{fig:postProb_rawScore_predClass_noDeg}
 \end{figure*}

In the above example we have purposefully chosen the coordinates of the test point $\bf{x}$ to be such that a degeneracy would occur, namely that the hyperplane along which the posterior probability is 0.5 (the fifth hyperplane here) classifies point $\bf{x}$ as negative, while the hyperplanes along which the posterior probabilities are 0.4 and 0.6 (the fourth and the sixth hyperplane here) classify point $\bf{x}$ as positive. Below we illustrate the procedure for estimating the (positive-class) posterior probability for a test point, which does not suffer from such a degeneracy. We follow again the four Steps outlined above, using the same nine hyperplanes, which correspond to posterior-probability estimates of $0.1, 0.2, \dots ,0.9$, and a new point $\bf{x}$, which is shown in Figure \ref{fig:9hp_pointX_noDeg}, along with the nine hyperplanes. 
\noindent Out of the 9 classifications for this new point $\bf{x}$, 7 are negative (from hyperplanes corresponding to expected posterior probabilities of 0.1, 0.2, 0.3, 0.4, 0.5, 0.6, and 0.7) and 2 are positive (from hyperplanes corresponding to expected posterior probabilities of 0.8 and 0.9) - see Figure \ref{fig:postProb_rawScore_predClass_noDeg}. Therefore the estimated positive-class posterior probability is $(2+1)/(9+2) = 3/11$. Note that this estimation, as already pointed out, is consistent with the strategy of estimating the class-posterior probability using a single hyperplane, which goes through point $\bf{x}$. This hyperplane, according to Figure \ref{fig:postProb_rawScore_predClass_noDeg}, should correspond to posterior-probability values between 1-0.7 and 1-0.8, as the classifications of the 9 hyperplanes change from negative to positive between locations for posterior-probabilities 0.7 and 0.8. Using sufficiently refined grid would yield the same estimate for both strategies in case of non-degeneracy for point $\bf{x}$. 

%
%

\section{Experiments on a UCI dataset}
\label{sec:UCIdataset}

\begin{figure*}[h]
 	\center
		\hbox{\hspace{-0.95cm} \includegraphics[height=13cm]{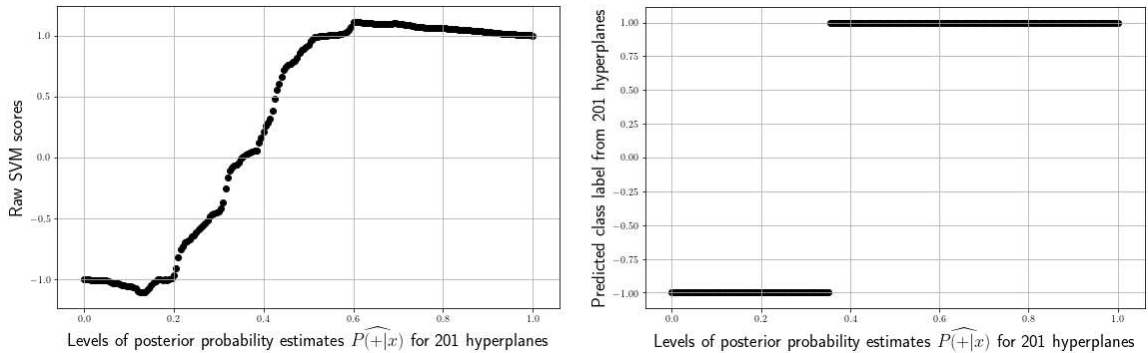}}
 	\vspace*{-7.5cm}
 	\caption{Raw SVM scores (left) and the corresponding predicted class labels (right) of the 201 hyperplanes for a point from the test set (original data: german-credit numeric dataset from UCI repository). The estimated positive-class posterior probability for the point is estimated as the relative number of positive-class classifications, in this case 65\%.}
 		\label{fig:postProb_rawScore_UCI}
 \end{figure*}

We have chosen a numeric dataset from the UCI repository \citep{UCI_Dua:2019}, the so-called ``german-credit'' dataset to illustrate the method of implied posterior probability estimation on a real dataset. We split the data into training and test set consecutively (rather than randomly, so that the results are reproducible), each set containing 500 instances. We follow the four estimation steps of implied posterior probability from Section \ref{sec:2Ddataset}. In Step 1 we build 201 models (hyperplanes \footnote{We use an SVM model with RBF kernel as a base (reference) SVM model, with parameters $C^+ = C^- = 10$, $z^+ = z^- = 0.5$ and $g = 0.001$, and vary the $z^+$ parameter between 0 and 1 with step 0.005, adjusting simultaneously the $z^-$ parameter to keep the effective number of points fixed. In this setting the implied posterior probability for the positive class is equal to $z^+$.}), such that the resulting estimated for the posterior probabilities are uniformly spaced in the interval [0,1], where the extremities of 0 and 1 have been defined by two fictitious models, resulting from applying weights of 0 and 1 for the positive class, respectively\footnote{We use the term fictitious, as the datasets in two cases contain only positive, and only negative instances, respectively} . We then iterate over all test points from the test dataset (Step 2). In Step 3 we classify each test point using all 201 SVM hyperplanes, and in Step 4 compute the final estimate for the positive-class posterior probability as the percentage of positive classifications by the 201 hyperplanes. For example, for the first test point the estimated (implied) posterior probability is 0.65, as illustrated in Figure \ref{fig:postProb_rawScore_UCI}, where 35\% of the hyperplanes classify the test point as negative class, and the rest 65\% as positive class. Note that there is no degeneracy in the estimation of the posterior probability in this case, as all hyperplanes that classify the test point as positive are associated with a posterior probability $>0.65$, and all hyperplanes that classify the test point as negative are associated with a posterior probability $<0.65$. This is illustrated in the figure by the fact that all classifications ``to the right'' of point 0.35 long the horizontal axis are positive.

\begin{figure*}[h]
 	\center
 		\includegraphics[height=12cm]{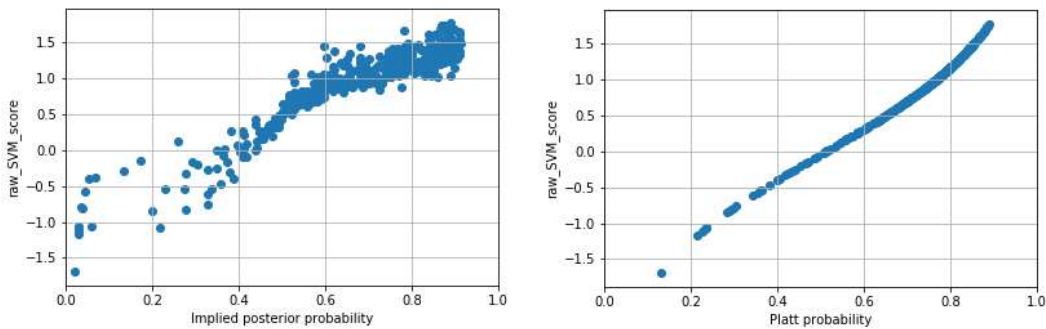}
 	\vspace*{-7cm}
 	\caption{Implied posterior probability estimates vs. raw SVM scores (left) and Platt posterior probability estimates vs. raw SVM scores (right) of the 201 hyperplanes for each point from the test set (original data: german-credit numeric dataset from UCI repository).}
 		\label{fig:implied_vs_raw_vs_platt}
 \end{figure*}

In Figure \ref{fig:implied_vs_raw_vs_platt} we show the relationship between raw SVM scores and estimated implied posterior probabilities over the test set, as well as the (popular alternative) Platt estimated posterior probabilities from raw SVM scores \citep{Platt99probabilisticoutputs}. The scores in the left panel of the figure are not based on one SVM model, and therefore need not be monotonically related to the original (single) SVM hyperplane built on the training dataset. We stress this due to the generally accepted requirement of monotonicity of the relationship between scores and posterior probability estimates, which are however produced by a single model and not by many models whose magnitudes of raw scores are not directly comparable, as in the case of implied posterior probability estimation. Thus, a raw score of, say, 3.1 produced by the $60^{th}$ hyperplane does not necessarily correspond to a raw score of 3.1 produced by the $70^{th}$ hyperplane. Or, alternatively, a point $\bf{x}$ with a score 3.1 from the $60^{th}$ SVM model would not have the same score from the $70^{th}$ SVM model.

Do the (final) estimates produced by the method of implied posterior probability approximate well the true posterior probabilities? This question can be answered by a calibration plot, where empirical posterior probabilities (on the test set) are compared to these estimates. This is usually achieved by splitting the ranges of the estimates into (equally-spaced) bins, and then computing the percentage of positive instances in each bin. If this percentage corresponds to the average estimated score for the posterior probabilities, we say that a good calibration exists, meaning that the estimated posterior probabilities correspond well to empirically derived posterior probabilities on a test set. Since the process of defining bin sizes is to a certain degree subjective, we will estimate the relation between scores and empirical posterior probability using the method of isotonic regression, which effectively determines the size of each bin automatically.

We point out that the method of implied posterior probabilities does not aim to provide estimates that are closer to the true posterior probabilities than other methods. Rather, the aim is to produce the most relevant scores pertaining to  (or, ``implied by'') the class of functions used to build a model on the data. These are the scores, loosely speaking, which the class of functions ``thinks'' should be the posterior probabilities, which are not necessarily the true posterior probabilities, especially so if this class of functions is very different from the underlying function which has generated the data. 

\begin{figure*}[h]
 	\center
		\hbox{\hspace{2.6cm} \includegraphics[height=17cm]{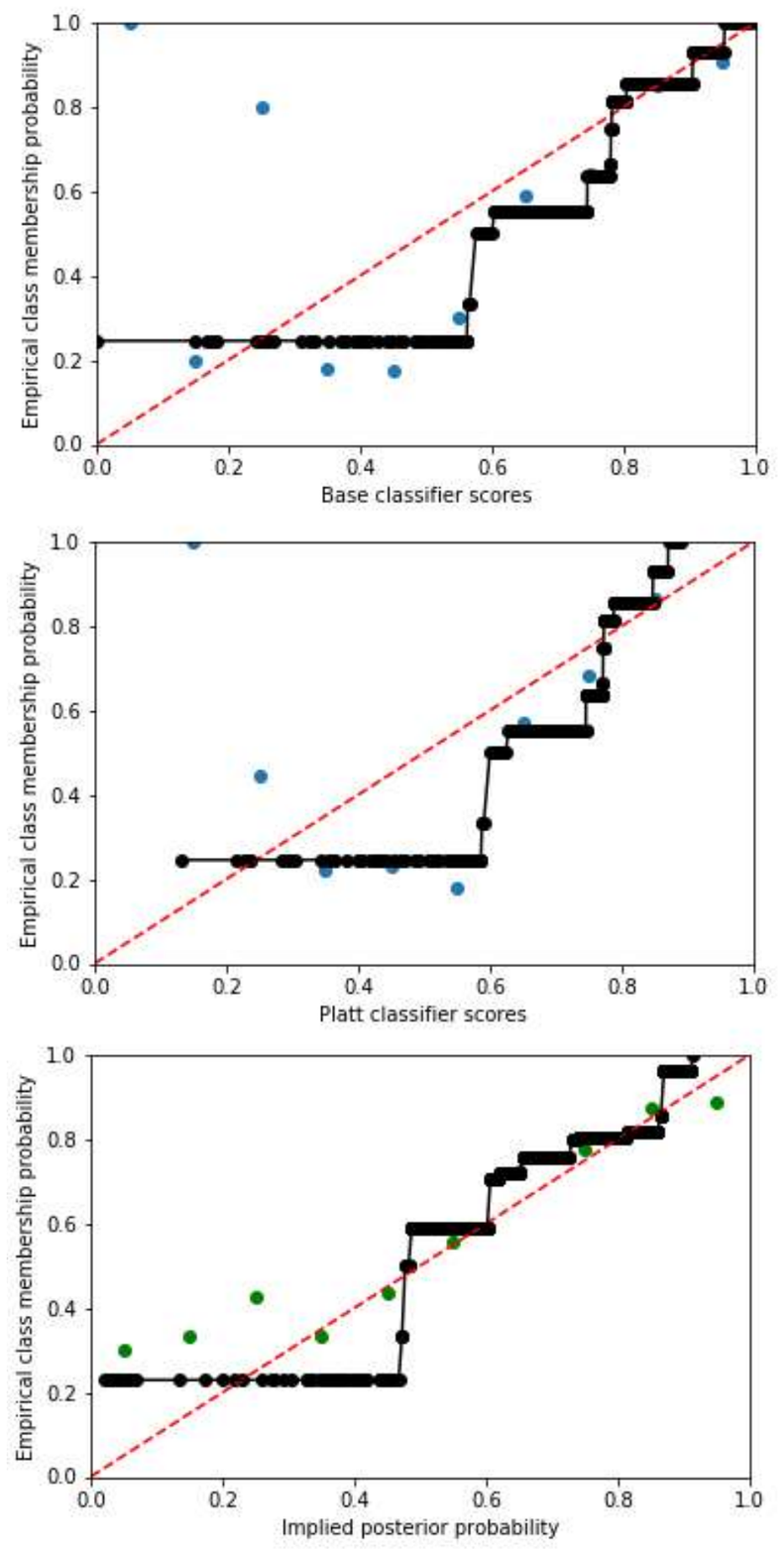}}
 	\vspace*{0cm}
 	\caption{Raw normalized SVM scores (upmost), Platt posterior probability estimates (middle), and implied posterior probability estimates (bottom) vs. isotonic regression out-of-sample estimates for posterior probabilities - plotted as step-function - as well as  bin-based estimates plotted as dots. The 45-degree line is shown as a reference to the best potential match. Data: a test set of 500 instances from the german-credit numeric dataset from the UCI repository.}
 		\label{fig:3iso}
 \end{figure*}

The calibration plots for 1) the relation between raw normalized (within the 0-1 range) SVM scores on the test set produced by the (single) SVM model built on the training dataset and empirical posterior probabilities, 2) the relation between Platt posterior probability scores on the test set produced by the (single) SVM model built on the training dataset and empirical posterior probabilities, as well as 3) the relation between implied posterior probability scores (produced by 201 SVM models) and empirical posterior probabilities, are shown in Figure \ref{fig:3iso}. For illustrative purposes, the dots in each subplot represent the expected empirical posterior probability over equally-spaced bins along the horizontal axes, where each bin contains a certain (non-equal) number of instances from the dataset with predicted score within the bin scoring range. In our case, we have used 10 bins, and therefore the (normalized) scores in each bin are 0-0.1, 0.1-0.2, etc. The isotonic regression step-functions for all three cases are shown as well. The inputs for the isotonic regression estimations are all individual scores (normalized SVM scores, Platt scores, or implied probability estimates) along with the realized classification over the test set.


\begin{figure*}[h]
 	\center
		\hbox{\hspace{0cm} \includegraphics[height=12cm]{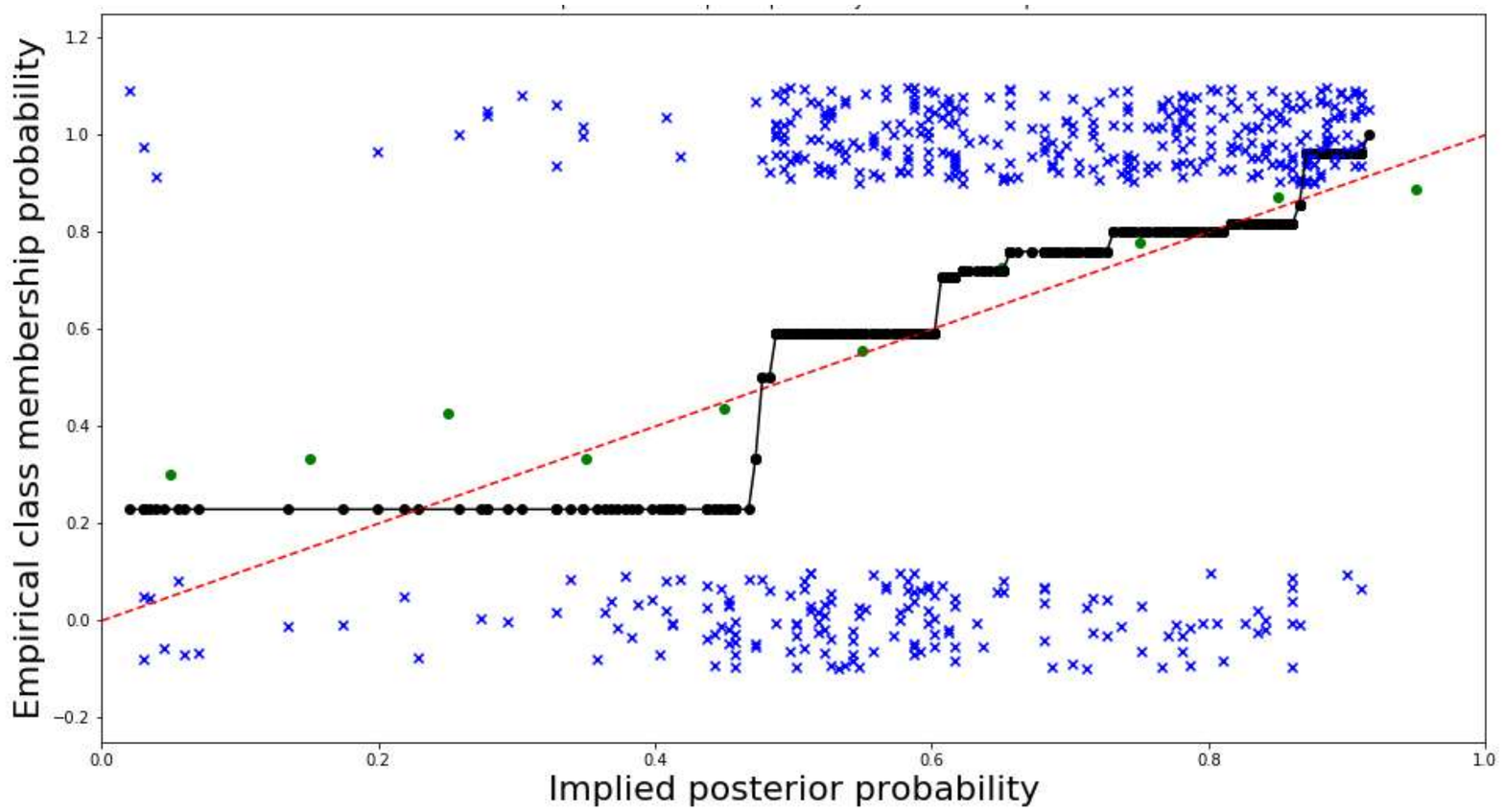}}
 	\vspace*{-3cm}
 	\caption{Implied posterior probability estimates vs. isotonic regression out-of-sample estimates for posterior probabilities - plotted as step-function (as in Figure \ref{fig:3iso}, bottom panel) - as well as 500 test-point class label estimates (jittered) plotted as crosses, used as input for the isotonic regression. The bin-based estimates are plotted as dots, where the bins are based on 10 equally spaces scores along the horizontal axis in the 0-1 interval.}
 		\label{fig:implied_iso_detailed}
 \end{figure*}

\begin{figure*}[h]
 	\center
		\hbox{\hspace{4cm} \includegraphics[height=6cm]{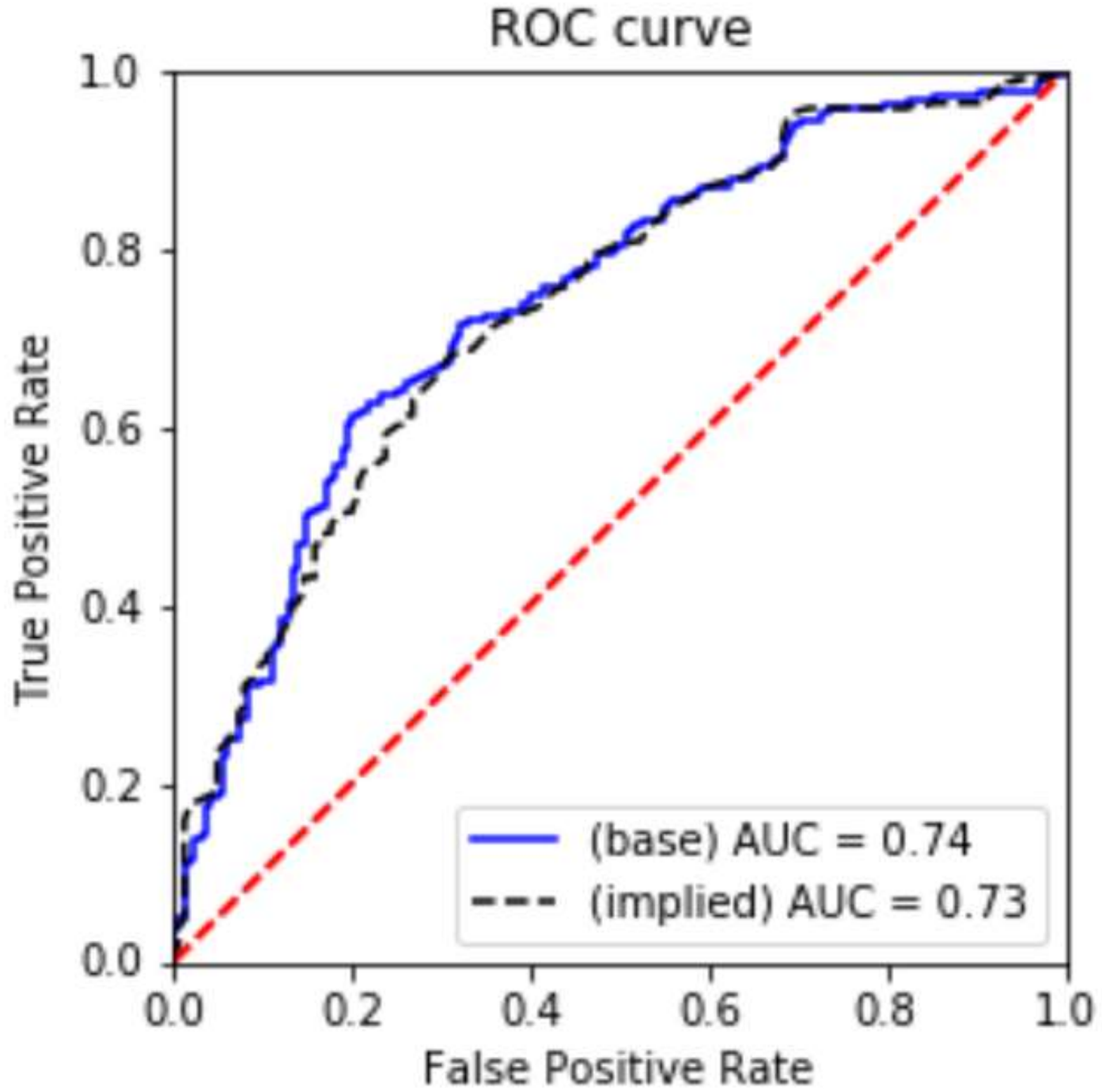}}
 	\vspace*{0cm}
 	\caption{ROC curves over the test set for the Platt scores (solid) and implied-probability estimates (dashed). The AUC values are 0.74 and 0.73 for the two curves, respectively.}
 		\label{fig:ROC_Platt_implied}
 \end{figure*}

The degree of calibration is accessed by a calibration score, defined here as the mean of the absolute differences between an estimated model-based posterior probability and the isotonic regression estimate for the posterior probability over all points from the test dataset. Thus, the calibration score for the reference SVM model is 0.085.  The calibration score for the Platt estimated posterior probabilities is 0.102. For the implied posterior probability estimation the corresponding calibration score is 0.065. Thus, on average the implied posterior probability score is 6.5\% (in absolute value) away from the estimated posterior probability value by the isotonic regression. In Figure \ref{fig:implied_iso_detailed} we show a detailed view of the input data for the isotonic regression, which consists of 500 pairs \{implied posterior probability estimate, observed class label\} from the test set. Some random noise has been added to the class label values for better visualization. In addition, we show the ROC curves associated with Platt estimates (solid line) and implied probability estimates (dashed line) over the test set in Figure \ref{fig:ROC_Platt_implied}. The AUC for the Platt-estimates case is 0.74, and the AUC for the implied-probabilities case is 0.73.

\section{Conclusion}
\label{sec:conclusion}

We have presented a detailed tutorial on the implied posterior probability estimation method applied to SVMs. The exposition is valid for any class of functions, where there is a natural way to change the relative weights of instances from classes in a dataset, as the case is for SVMs. Even if there is no such natural way, then one option would be to use random down-sampling and up-sampling of the classes. We have discussed the binary classification task only, and leave the multi-class treatment for future research.

\vskip 0.2in
\bibliography{postprob_SVM_tutorial}

\end{document}